\documentclass[conference]{IEEEtran}
\IEEEoverridecommandlockouts
\usepackage{cite}
\usepackage{amsmath,amssymb,amsfonts}
\usepackage{algorithmic}
\usepackage{graphicx}
\usepackage{textcomp}
\usepackage{xcolor}
\usepackage[bookmarks=false]{hyperref}

\usepackage{float}

\usepackage[group-separator={,}]{siunitx}
\def\BibTeX{{\rm B\kern-.05em{\sc i\kern-.025em b}\kern-.08em
    T\kern-.1667em\lower.7ex\hbox{E}\kern-.125emX}}

\usepackage{booktabs} 

\definecolor{red}{RGB}{255,0,0}
\definecolor{green}{RGB}{0, 102, 0}
\definecolor{maron}{RGB}{102,51,0}
\definecolor{orange}{RGB}{255,128,0}
\definecolor{blue}{RGB}{0,153,153}

\newcommand{\fasttext}{\textbf{fastText}%
}
\newcommand{\bpemb}{\textbf{BPEmb}}
\newcommand{\fasttextatt}{\textbf{fastTextAtt}%
}
\newcommand{\bpembatt}{\textbf{BPEmbAtt}}
\newcommand{\fasttextadann}{\textbf{fastTextADANN}}
\newcommand{\bpembadann}{\textbf{BPEmbADANN}}
\newcommand{\fasttextadannnoisy}{\textbf{fastTextADANNNoisy}}
\newcommand{\bpembadannnoisy}{\textbf{BPEmbADANNNoisy}}

\begin{document}
\title{Multinational Address Parsing: A Zero-Shot Evaluation
}

\author{\IEEEauthorblockN{Marouane Yassine, David Beauchemin, François Laviolette, Luc Lamontagne}
\IEEEauthorblockA{\textit{Department of Computer Science and Software Engineering, Laval University} \\
\textit{Group for Research in Artificial Intelligence of Laval University (GRAIL)}, \\
Québec, Canada \\
marouane.yassine.1@ulaval.ca, david.beauchemin.5@ulaval.ca, francois.laviolette@ift.ulaval.ca, luc.lamontagne@ift.ulaval.ca}
}

\maketitle

\begin{abstract}
Address parsing consists of identifying the segments that make up an address, such as a street name or a postal code. Because of its importance for tasks like record linkage, address parsing has been approached with many techniques, the latest relying on neural networks. While these models yield notable results, previous work on neural networks has only focused on parsing addresses from a single source country. 
This paper explores the possibility of transferring the address parsing knowledge acquired by training deep learning models on some countries' addresses to others with no further training in a zero-shot transfer learning setting. We also experiment using an attention mechanism and a domain adversarial training algorithm in the same zero-shot transfer setting to improve performance. Both methods yield state-of-the-art performance for most of the tested countries while giving good results to the remaining countries. We also explore the effect of incomplete addresses on our best model, and we evaluate the impact of using incomplete addresses during training. In addition, we propose an open-source Python implementation of some of our trained models\footnote{\href{https://github.com/GRAAL-Research/deepparse}{\url{https://github.com/GRAAL-Research/deepparse}}}.
\end{abstract}


\section{Introduction}
\emph{Address Parsing} is the task of decomposing an address into the different components it is made of. This task is an essential part of many applications, such as geocoding and record linkage. Indeed, to find a particular location based on textual data, it is quite useful to detect the different parts of an address to make an informed decision. Similarly, comparing two addresses to decide whether two or more database entries refer to the same entity can prove to be quite difficult and prone to errors if based on methods such as edit distance algorithms given the various address writing standards.

There have been many efforts to solve the address parsing problem. From rule-based techniques \cite{rule-based} to probabilistic approaches and neural network models \cite{8615844}, a lot of progress has been made in reaching an accurate segmentation of addresses. These previous pieces of work did a remarkable job at finding solutions for the challenges related to the address parsing task. However, most of these approaches either do not take into account parsing addresses from different countries or do so but at the cost of a considerable amount of meta-data and elaborate data pre-processing pipelines \cite{rnn-parsing, hmm-parsing, crf-parsing, feedforward-parsing}. 

Our work comes with three contributions. First, we propose an approach for multinational address parsing using a Recurrent Neural Network (RNN) architecture by building on, and improving, our previous work on the subject \cite{9357170}. Secondly, we evaluate the degree to which the improved models trained on countries' addresses data can perform well at parsing addresses from other countries. Finally, we evaluate the performance of our models on incomplete addresses and propose a method to improve their accuracy.
The article outline is as follows: we present the related work in \autoref{sec:relatedwork}, our proposed architectures and approaches in \autoref{sec:architectures}, the datasets that we used for our experiments in \autoref{sec:data}, our experimental settings in \autoref{sec:exp}, our results on complete addresses in \autoref{sec:completeresults} and on addresses with missing values in \autoref{sec:missingresults}, and finally we conclude our article with an overview of our work and future work in \autoref{sec:conclusion}.

\section{Related Work}
\label{sec:relatedwork}

Since address parsing is a sequence tagging task, it has been tackled using probabilistic methods mainly based on Hidden Markov Models (HMM) and Conditional Random Fields (CRF) \cite{hmm-parsing, crf-parsing, 8615844}. For instance, \cite{hmm-parsing} proposed a large scale HMM-based parsing technique capable of segmenting a large number of addresses whilst being robust to possible irregularities in the input data. In addition, \cite{crf-parsing} implemented a discriminative model using a linear-chain CRF coupled with a learned Stochastic Regular Grammar (SRG). This approach allowed the authors to better address the complexity of the features while capturing higher-level dependencies by applying the SRG on the CRF outputs as a score function, thus taking into account the possible lack of features for a particular token in a lexicon-based model. These probabilistic methods usually rely on structured data as well as some sort of prior knowledge of this data for feature extraction or in order to implement algorithms such as Viterbi \cite{viterbi}, especially in the case of generative methods. 

In recent years, new methods \cite{rnn-parsing, 8615844} utilizing the power of neural networks have been proposed as solutions for the address parsing problem. Using a single hidden layer feed-forward model, \cite{feedforward-parsing} achieved good performance. However, their approach relied on a pre-processing and post-processing pipeline to deal with the different structures of address writing and the possible prediction errors. For instance, the input data is normalized to reduce noise and standardize the many variations that can refer to the same word, such as \emph{road} and \emph{rd}. In addition, the model's predictions are put through a rule-based validation step to make sure that they fit known patterns. In contrast, \cite{rnn-parsing} proposed a deep learning approach based on the use of RNN and did not use any pre or post-processing. Their experiments focused on comparing the performance of both unidirectional and bidirectional vanilla RNN and Long-Short Term Memory Models (LSTM) \cite{LSTM}, as well as a Seq2Seq. The models achieved high accuracy on test sets with the Seq2Seq leading the scoreboard on most of them with no particular pre-processing needed during the inference process.

Despite reaching notable performances, the aforementioned approaches are limited to parsing addresses from a single country and would need to be adjusted to support a multinational scope of address parsing. To tackle this problem, Libpostal\footnote{\href{https://github.com/openvenues/libpostal}{\url{https://github.com/openvenues/libpostal}}}, a library for international address parsing, has been proposed. This library uses a CRF-based model trained with an averaged Perceptron for scalability. The model was trained on data from each country in the world and was able to achieve a $99.45~\%$ full parse accuracy\footnote{The accuracy was computed considering the entire sequence and was not focused on individual tokens.}, thus defining a new state-of-the-art\footnote{For a comparison of some of our models with Libpostal, visit our previous article \cite{9357170}.}. However, this requires putting addresses through a heavy pre-processing pipeline before feeding them to the prediction model. 

Taking into account that some countries' addresses share the same language and format (i.e: the order of the address elements), \cite{9357170} have proposed to leverage subword embeddings with a Seq2Seq architecture to achieve multinational address parsing. They achieved a $\mathbf{99~\%}$ parse accuracy on 20 countries. They also proposed to explore whether their trained model had learned transferable representation in a zero-shot evaluation procedure. Finally, they tested the parsing accuracy on 41 countries not seen during training and achieved at least $\mathbf{80~\%}$ parse accuracy for most of these countries.

This work aims to enhance the performance of a single model capable of parsing addresses from multiple countries, to explore the possibility of zero-shot transfer from some countries' addresses to others' and to explore (plus improve) the performance on incomplete addresses. In this respect, attention mechanisms have become ubiquitous when it comes to improving the performance of natural language processing models. These mechanisms come in different formats \cite{attn_nlp} and have been applied to various tasks such as Machine Translation \cite{luong-etal-2015-effective} Named Entity Recognition \cite{attn_ner, attn_ner_2} and Sentiment Analysis \cite{wang-etal-2016-attention}. Moreover, the Transformer architecture \cite{NIPS2017_3f5ee243} which is based on self-attention has become the state-of-the-art in an overwhelming majority of natural language processing tasks. Zero-shot transfer has also been applied in similar settings to ours, particularly in cross lingual transfer. For instance, \cite{zero-shot-bert} propose a linear transformation which projects the contextual representations obtained by BERT \cite{devlin-etal-2019-bert} for certain target languages into the space of a source language, on which the model was trained to achieve the task of dependency parsing. This strategy managed to improve the model's performance in zero-shot cross lingual dependency parsing. As for \cite{DBLP:conf/aaai/JiZDZCL20}, they have used cross-lingual pre-training of a Transformer encoder in order to achieve zero-shot machine translation for source-target language pairs with scarce parallel training data. Another technique that has proven to be effective in the context of zero-shot transfer is domain adaptation. In this regard, \cite{peng2018zeroshot} and \cite{9447890} approach the problem of zero-shot domain adaptation, in which target domain data for a particular task is completely absent, by training a model on an irrelevant task for which data from both source and target domains is available, while also utilizing data from the source domain to learn the objective task. We leverage both attention mechanisms and domain adaptation in an attempt to build a model capable of performing well on the task of address parsing for training countries and transferring the obtained knowledge to other countries without further training.

\section{Architectures}
\label{sec:architectures}
Our models' architectures are based on previous work conducted in \cite{9357170}. They are composed of an embedding model and a tagging model. The base architectures are improved using an attention mechanism as well as domain adaptation. These improvements are discussed in Subsections \ref{subsec:attention} and \ref{subsec:adann} respectively. Subsection~\ref{subsec:base} introduces the base architecture.

\subsection{Base Architecture}
\label{subsec:base}

\subsubsection{Embedding Model}
\label{subsec:embedding}
An embedding model is essential in deep learning architectures solving natural language processing problems since it converts textual data into vector representations which are then used as inputs to the downstream models. Since we are dealing with a multilingual dataset, the used embedding model must be able to handle multiple languages.

Firstly, we use a fixed pre-trained monolingual fastText model \cite{fasttext} (pre-trained on the French language). The French embeddings were chosen since the French language shares Latin roots with many languages in our test set. It is also due to the considerable size of the corpus on which these embeddings were trained. Despite its monolingual characteristic, this embedding model is able to generate vector representations for words outside its training vocabulary by utilizing subword information in the form of character n-grams. We refer to this embedding model as FastText.

Secondly, we use an encoding of words using multilingual subword embeddings provided by MultiBPEmb \cite{bpemb}. This model relies on subword units to produce vector representations, therefore producing one or more vectors for each word. Consequently, we merge the obtained embeddings into a single word embedding using a Bidirectional LSTM (Bi-LSTM) with a hidden state dimension of 300. We build the word embeddings by running the concatenated forward and backward hidden states corresponding to the last time step for each word decomposition through a fully connected layer of which the number of neurons is equal to the dimension of the hidden states. This approach produces 300-dimensional word embeddings. We refer to these embeddings as BPEmb.

\subsubsection{Tagging Model}
\label{subsec:tagging}

The used downstream tagging model is a Seq2Seq model consisting of a one-layer unidirectional LSTM encoder and a one-layer unidirectional LSTM decoder followed by a fully-connected linear layer with a softmax activation. Both the encoder's and decoder's hidden states are of dimension $1024$. The embedded address sequence is fed to the encoder that produces hidden states, the last of which is used as a context vector to initialize the decoder's hidden states. The decoder is then given a Beginning Of Sequence (BOS) token as input and, at each time step, the prediction from the last step is used as input. To better adapt the model to the task of sequence labeling and to facilitate the convergence process, we only require the decoder to produce a sequence with the same length as the input address. The decoder's outputs are forwarded to the linear layer of which the number of neurons is equal to the tag space dimensionality. The softmax activation function computes probabilities over the linear layer's outputs to predict the most likely token at each time step.

\subsection{Attention Architecture}
\label{subsec:attention}
Attention mechanisms are neural network components that can produce a distribution describing the interdependence between a model's inputs and outputs (general attention) or amongst model inputs themselves (self-attention). These mechanisms are common in natural language processing encoder-decoder architectures such as neural machine translation models \cite{nmt} since they have shown to improve models' performance and help address some of the issues recurrent neural networks suffer from when it comes to dealing with long sequences.

Attention mechanisms are also exploited for the interpretability of neural networks, where they are considered to provide insights about the impact of some neural network's inputs on its predictions. This use of attention mechanisms has been contested in \cite{jain-wallace-2019-attention} because of a lack of consistency with feature importance measures, among other things. However, other work has suggested that attention mechanisms provide a certain degree of interpretability depending on the task at hand \cite{wiegreffe2019attention} \cite{vashishth2019attention}. In this work, we focus on the performance enhancement of address tagging models using attention.

\subsubsection{Attention models}
The implementation of our attention models was inspired by \cite{nmt}. The models' architecture remains similar to that of our base models, with some alterations to the decoding process. Indeed, instead of feeding the last predicted tag as an input to the decoder at the current time step $i$, we compute an input using the encoder's outputs $\vec{O}$, the last decoder hidden state $h_{i-1}$, and the last predicted tag's representation $\vec{t}$. 

We start by computing attention weights as follows:

\begin{center}
$ \displaystyle \alpha_{i, j} = \frac{\exp({a_{i, j})}}{\sum_{k} \exp({a_{i, k}})}$
\end{center}

where

\begin{center}
$ a_{i, j} = p \times tanh(W_{h} h_{i-1} + W_{o} O_{j})$
\end{center}

and $W_{h}$, $W_{o}$ and $p$ are learnable parameters.

Next, we compute a context vector by weighting the encoder's outputs with the obtained attention weights:

\begin{center}
    $\vec{c_{i}} = \sum_{k} \alpha_{i, k} O_{k}$
\end{center}

Finally, we obtain the decoder's input by concatenating the last prediction $\vec{t}$ with the context vector $\vec{c_{i}}$.
Note that the first decoder's input is computed using the last encoder's hidden state.

We use this approach with the two aforementioned embedding methods and name the obtained models \fasttextatt{} and \bpembatt{} respectively.

\subsection{Domain Adaptation}
\label{subsec:adann}
Domain adaptation is a branch of transfer learning which aims at applying a model trained on data from a source domain to data from a target domain which somewhat differs from one another but still retains a certain degree of similarity. More specifically, it is a technique used when the input and output belong to the same space, but the probability distribution which associates them changes as we move from one domain to the other \cite{da-tl}. Our objective is to generalize the performance of address parsing models to countries of which no data is used at training time. In order to extend our models to cope with this domain adaptation problem, we enhance our base approach using domain adversarial neural networks. 

\subsubsection{Domain Adversarial Neural Networks}
Domain adversarial neural networks \cite{dann} achieve domain adaptation by appending a second parallel output layer to a neural network classifier, the purpose of which is to predict the domain of the network's input. Since the target domain labels are not available during training, two losses are computed when the input belongs to the source domain (i.e. the labels classification loss and the domain classification loss), whilst only the domain classification loss is computed when the input belongs to the target domain. Moreover, the gradient associated with the domain classification layer is reversed during backpropagation. This aims to hinder the neural network's ability to differentiate between the source and target domains while still learning to perform well on classifying data from the source domain.

\subsubsection{Implementation}
First of all, we modified our neural network architecture by adding a domain discriminator in the form of a fully connected layer with two output neurons which takes the context vector produced by the Seq2Seq encoder as input. This layer is preceded by a gradient reversal layer that reverses the computed gradient sign during backpropagation. The domain discriminator is followed by a Softmax activation function, and its loss is computed using a Cross-Entropy loss function. Secondly, we used the ADANN \cite{adann} training algorithm to train our model since it is designed to enable multi-domain adversarial training by considering, during the forward pass of each batch, one domain as a source domain and another random domain as a target domain, and so on for each of the available domains; our domains being the countries for which training data is available. We hope, by using this approach, to construct models that can learn to parse addresses from countries with different address formats without making a significant distinction between them. Therefore these models would perform better on parsing addresses from countries not seen during their training.

We use this approach with the two aforementioned embedding methods and name the obtained models \fasttextadann{} and \bpembadann{} respectively.

\section{Data}
\label{sec:data}
\subsection{Complete Address Dataset}

We use the same dataset as per \cite{9357170}\footnote{\href{https://github.com/GRAAL-Research/deepparse-address-data}{\url{https://github.com/GRAAL-Research/deepparse-address-data}}}, Deepparse dataset.
Which is built on the open-source dataset on which Libpostal's models were trained.
Libpostal dataset is based on three open source addresses datasets: OpenStreetMap, Yahoo GeoPlanet and OpenAddresses. 
It contains over 1 billion addresses from nearly 241 countries \cite{libpostal}.
Using this dataset, \cite{9357170} have selected 61 countries based on twofold: availability of at least 500 addresses, availability of the addresses components type.
The address data of those 61 countries are in their official language.
For example, Korean addresses are in Korean while Canada's addresses are in the two official languages, French and English. 
Among the 61 selected countries, twenty countries are used for multinational training, and forty-one are used for zero-shot transfer evaluation. 
The dataset uses the following eight tags: StreetNumber, StreetName, Unit, Municipality, Province, PostalCode, Orientation, and GeneralDelivery.

\subsection{Incomplete Address Dataset}
\label{subsec:iad}
We introduce a second dataset based on Deepparse dataset. It is similar to the Complete address dataset. 
It is composed of the same twenty countries used for multinational training but with a sample size of \num{50000} for training and \num{25000} for holdout evaluation. 
The dataset consists only of addresses where each one is missing at least one of the following four tags: StreetName, PostalCode, Municipality and Province. We consider an address incomplete if it is not composed of at least all of the four tags. 
For example, the sequence of tags for the address ``221 B Baker Street" is \{StreetNumber, Unit, StreetName, StreetName\}, and it is incomplete since the PostalCode and the Municipality tags are missing. We will refer to this dataset as the ``incomplete address" dataset.

\section{Experiments}
\label{sec:exp}
For our experiments, as per \cite{9357170} we trained each of our four models (\fasttextatt{}, \fasttextadann{}, \bpembatt{} and \bpembadann{}) five times\footnote{Using each of the following seeds $\{5, 10, 15, 20, 25\}$. When a model did not converge (a high loss value on train and validation), we retrain the model using a different seed (30).} for 200 epochs with a batch size of \num{512} for the base approach and the attention models and \num{256} for the ADANN one. An early stopping with a patience of fifteen epochs was also applied during training. We initialize the learning rate at 0.1 and use learning rate scheduling to lower it by a factor of 0.1 after ten epochs without loss reduction. Our loss function of choice is the Cross-Entropy loss due to its suitability for the softmax function. The optimization is done through Stochastic Gradient Descent.

As per \cite{9357170}, we use teacher forcing \cite{6795228} to speed up the convergence. The architecture and the training of the models were implemented using Pytorch \cite{pytorch}, and Poutyne \cite{poutyne}.

\subsection{Evaluation Procedure}
We train our four models on our multinational dataset, the difference between the models being (1) the word embedding method employed (fastText and BPEmb) and (2) the use of attention mechanism or domain adaption learning. Each model has been trained five times, and we report the models' mean accuracy and standard deviation on the per-country zero-shot data. The accuracy for each sequence is computed as the proportion of the tags predicted correctly by the model. As such, predicting all the tags for a sequence correctly yields a perfect accuracy. More precisely, errors in tag predictions have an impact on the accuracy for a given sequence. However, the accuracy will not be null unless all the predicted tags for the sequence are incorrect. These results will be discussed in \autoref{sec:completeresults}.

\subsection{Incomplete Address Evaluation Procedure}
Since addresses do not always include all the components, we also evaluate our four models on the incomplete addresses dataset introduced in the Subsection~\ref{subsec:iad}. We hypothesize that an incomplete address can confuse our models since we use a Seq2Seq architecture, and the compressed representation of an incomplete address will not be the same as the same complete one. For example, the address ``221 B Baker Street London NW1 6XE" is complete and is a typical way to write an address. But, many addresses are not always in such a form. Such as the address ``221 B Baker Street", which is the same as the previous one but without the city and the postal code. That difference can be more challenging for our models since postal code is usually a good way to tell the difference between the pattern. We will also evaluate the performance of two new models, \fasttextadannnoisy{} and \bpembadannnoisy{}), trained on the complete and incomplete addresses dataset to investigate if the addition of incomplete address help improves performance on that type of data. These results will be discussed in \autoref{sec:missingresults}.

\section{Complete addresses Results}
\label{sec:completeresults}
In this section, we present and discuss the results of all our four trained models plus the two models in \cite{9357170} (\fasttext{} and \bpemb{}). We first evaluated them on the holdout addresses dataset and the zero-shot addresses dataset.

\subsection{Multinational Evaluation}
\label{subsec:multieval}
\begin{table*}
    \centering
    \caption{Mean accuracy and standard deviation obtained with multinational models on the holdout dataset for training countries. \textbf{bold} values indicate the best accuracy for each country.}
    \label{table:multinational-holdout}
   \resizebox{0.85\textwidth}{!}{%
    \begin{tabular}{lrrrrrr}
        \toprule
        Country & FastText & BPEmb & FastTextAtt & BPEmbAtt & FastTextADANN & BPEmbADANN\\
        \midrule
        United States & $99.61 \pm 0.09$ & $99.67 \pm 0.09$ & $\mathbf{99.73 \pm 0.02}$ & $99.65 \pm 0.23$ & $99.70 \pm 0.03$ & $99.68 \pm 0.16$\\
        Brazil & $99.40 \pm 0.10$ & $99.42 \pm 0.15$ & $\mathbf{99.58 \pm 0.04}$ & $99.42 \pm 0.39$ & $99.53 \pm 0.04$ & $99.42 \pm 0.34$\\
        South Korea & $99.96 \pm 0.01$ & $\mathbf{100.00 \pm 0.00}$ & $99.98 \pm 0.01$ & $\mathbf{100.00 \pm 0.00}$ & $99.98 \pm 0.01$ & $\mathbf{100.00 \pm 0.00}$\\
        Australia & $99.68 \pm 0.05$ & $\mathbf{99.80 \pm 0.05}$ & $99.77 \pm 0.03$ & $99.78 \pm 0.13$ & $99.76 \pm 0.03$ & $99.77 \pm 0.17$\\
        Mexico & $99.60 \pm 0.06$ & $99.68 \pm 0.06$ & $\mathbf{99.71 \pm 0.03}$ & $99.70 \pm 0.14$ & $99.68 \pm 0.02$ & $99.69 \pm 0.12$\\
        Germany & $99.77 \pm 0.04$ & $99.89 \pm 0.03$ & $99.85 \pm 0.02$ & $99.90 \pm 0.08$ & $99.84 \pm 0.01$ & $\mathbf{99.91 \pm 0.05}$\\
        Spain & $99.75 \pm 0.05$ & $99.85 \pm 0.04$ & $99.83 \pm 0.02$ & $\mathbf{99.86 \pm 0.09}$ & $99.80 \pm 0.02$ & $99.83 \pm 0.11$\\
        Netherlands & $99.61 \pm 0.07$ & $99.88 \pm 0.03$ & $99.75 \pm 0.03$ & $99.90 \pm 0.07$ & $99.72 \pm 0.02$ & $\mathbf{99.91 \pm 0.05}$\\
        Canada & $99.79 \pm 0.05$ & $\mathbf{99.87 \pm 0.04}$ & $99.87 \pm 0.02$ & $99.87 \pm 0.10$ & $99.85 \pm 0.01$ & $99.86 \pm 0.10$\\
        Switzerland & $99.53 \pm 0.09$ & $99.75 \pm 0.08$ & $99.62 \pm 0.05$ & $99.77 \pm 0.14$ & $99.59 \pm 0.05$ & $\mathbf{99.82 \pm 0.12}$\\
        Poland & $99.69 \pm 0.07$ & $99.89 \pm 0.04$ & $99.80 \pm 0.02$ & $99.90 \pm 0.07$ & $99.78 \pm 0.02$ & $\mathbf{99.92 \pm 0.04}$\\
        Norway & $99.46 \pm 0.06$ & $98.41 \pm 0.63$ & $99.44 \pm 0.11$ & $98.20 \pm 1.13$ & $\mathbf{99.53 \pm 0.04}$ & $97.95 \pm 0.44$\\
        Austria & $99.28 \pm 0.03$ & $98.98 \pm 0.22$ & $\mathbf{99.38 \pm 0.06}$ & $98.96 \pm 0.37$ & $99.30 \pm 0.07$ & $99.34 \pm 0.32$\\
        Finland & $99.77 \pm 0.03$ & $\mathbf{99.87 \pm 0.01}$ & $99.83 \pm 0.02$ & $99.86 \pm 0.01$ & $99.82 \pm 0.01$ & $99.84 \pm 0.01$\\
        Denmark & $99.71 \pm 0.07$ & $99.90 \pm 0.03$ & $99.82 \pm 0.03$ & $99.91 \pm 0.06$ & $99.80 \pm 0.02$ & $\mathbf{99.93 \pm 0.05}$\\
        Czechia & $99.57 \pm 0.09$ & $99.89 \pm 0.04$ & $99.73 \pm 0.02$ & $99.89 \pm 0.10$ & $99.70 \pm 0.02$ & $\mathbf{99.90 \pm 0.06}$\\
        Italy & $99.73 \pm 0.05$ & $99.81 \pm 0.05$ & $99.83 \pm 0.02$ & $\mathbf{99.83 \pm 0.11}$ & $99.80 \pm 0.02$ & $99.82 \pm 0.11$\\
        France & $99.66 \pm 0.08$ & $99.69 \pm 0.11$ & $\mathbf{99.79 \pm 0.04}$ & $99.69 \pm 0.22$ & $99.77 \pm 0.03$ & $99.70 \pm 0.17$\\
        UK & $99.61 \pm 0.10$ & $99.74 \pm 0.08$ & $\mathbf{99.77 \pm 0.05}$ & $99.72 \pm 0.20$ & $99.73 \pm 0.03$ & $99.72 \pm 0.20$\\
        Russia & $99.03 \pm 0.24$ & $\mathbf{99.67 \pm 0.11}$ & $99.40 \pm 0.13$ & $99.54 \pm 0.39$ & $99.23 \pm 0.12$ & $99.59 \pm 0.31$\\
        \midrule
        Mean & $99.61 \pm 0.20$ & $99.68 \pm 0.36$ & $\mathbf{99.72 \pm 0.16}$ & $99.67 \pm 0.40$ & $99.70 \pm 0.18$ & $99.68 \pm 0.43$\\
        \bottomrule
    \end{tabular}%
    }
\end{table*}

\autoref{table:multinational-holdout} presents all the models' mean accuracy and standard deviation on the holdout dataset for training countries. First, we find that South Korea is the only country where a perfect accuracy was achieved when using byte-pairs embeddings (BPemb) or almost all the time (four seeds out of five) when using fastText embeddings. Since South Korea is the only country using a different pattern in the training set where the province and municipality occur before the street name, it seems that our models might have memorized this particular pattern. To validate this intuition, we randomly reordered \num{6000} South Korean addresses to follow either the first (red) or the second (brown) address pattern (equally divided between the two). We observe, after this reordering, that the mean accuracy drops to $28.04~\%$ considering that using a random tags annotation, we get a $12.29~\%$ accuracy.

It is also interesting to notice that the models' accuracies are good when using fastText monolingual word embeddings, especially on South Korean addresses despite the entirely different alphabet. These results illustrate that our model, regardless of the embeddings model, learned the representation of an address sequence even if the words' representations are not native to the language (French vs Korean). 

Finally, all our models achieve state-of-the-art performance on our holdout dataset while using less data than previous approaches (e.g. Libpostal) and neither pre nor post-processing. However, at this point, it is difficult to conclude which of our models is the leading one. In the following subsection, we investigate the zero-shot performance of our models on countries not seen during training.

\subsection{Zero-Shot Evaluation}
Since training a deep learning model to parse addresses from every country in the world would require a significant amount of data and resources, our ongoing work aims at achieving domain adaptation to be able to train on a reasonable amount of data and generalize to data from different sources. We begin by exploring how well our architecture can generalize in a zero-shot manner. To this end, we test each of our four models on address data from countries not seen during the training. The results are reported in \autoref{table:zero-shot-results}.

\subsubsection{Attention Models}
In an attempt to improve our models' zero-shot performance, the base architecture was augmented with an attention mechanism as described in Subsection~\ref{subsec:attention}. The results of those two models (\fasttextatt{} and \bpembatt{}) are compared to their base approach equivalents respectively in \autoref{table:zero-shot-results}. For the model \fasttextatt{} (left section of the table), we observe that attention mechanisms improve the performance for 20 out of 41 countries by more than $1~\%$. Where 10 out of 41 improves with more than $2~\%$. Also, 10 out of 41 countries' accuracies were increased by less than $1~\%$. However, for the other countries (11 out of 41), we observe that the accuracy is less than $0.5~\%$ poorer than the base approach for most of them. The attention mechanism improved two countries' accuracies above the $90~\%$ accuracy (Belgium and Lithuania) and one above the $80~\%$ (Philippines). \textbf{FastTextAtt} achieves good results for $83~\%$ of the countries (34), almost $52~\%$ of which (21) are near state-of-the-art performance. 

In contrast, the results for the \bpembatt{} model (right section of the table) are not as good. We observe that results were be increased by at least more than $1~\%$ for 14 out of 41 countries. While 11 out of 41 saw improvement by more than $2~\%$. Also, 9 out of 41 countries improved by less than $1~\%$. However, for the other countries (20), we observe that the accuracy is less than $0.5~\%$ poorer than the base approach for half of them. For the other half, results are between $1~\%$ and $2~\%$ lower, thus lowering performance for two countries (India and Bangladesh) below $80~\%$ and one (Malaysia) below $90~\%$. Hence, the use of attention mechanism with byte-pair multilingual embeddings lowers performance overall, especially since only one country (Estonia) yields results above the $80~\%$ compared to the base approach. \bpembatt{} achieves good results for $78~\%$ of the countries (32), almost $50~\%$ of which (20) is near state-of-the-art performance. 

Finally, we observe that in some cases, the use of an attention mechanism can substantially improve performance, such as Greece where the increase is $4~\%$ for \fasttextatt{} and near $16~\%$ for \bpembatt{}. We also observe a smaller variance for both the models using attention mechanisms, meaning those models are more stable during training and converge to a better optimum. Overall, these results show that our attention mechanisms architecture can generally yield better accuracies during training.

\begin{table*}
    \centering
    \caption{Mean accuracy (and standard deviation) per country for zero-shot transfer models - base approach versus attention models. \textbf{bold} values indicate the best accuracy for each embeddings type.}
    \label{table:zero-shot-results}
    \resizebox{\textwidth}{!}{%
    \begin{tabular}{lrr|rrlrr|rr}
    \toprule
    Country & FastText & FastTextAtt & BPEmb & BPEmbAtt & Country & FastText & FastTextAtt & BPEmb & BPEmbAtt\\
    \midrule
    Belgium & $88.14 \pm 1.04$ & $\mathbf{90.01 \pm 1.59}$ & $87.29 \pm 1.40$ & $\mathbf{89.35 \pm 0.48}$ & Faroe Islands & $\mathbf{74.14 \pm 1.83}$ & $74.00 \pm 1.00$ & $85.50 \pm 0.11$ & $\mathbf{85.89 \pm 1.08}$\\
    Sweden & $81.59 \pm 4.53$ & $\mathbf{85.73 \pm 2.30}$ & $\mathbf{90.76 \pm 3.03}$ & $90.64 \pm 5.07$ & Réunion & $96.80 \pm 0.45$ & $\mathbf{96.86 \pm 1.27}$ & $93.67 \pm 0.26$ & $\mathbf{94.19 \pm 0.27}$\\
    Argentina & $86.26 \pm 0.47$ & $\mathbf{87.76 \pm 0.71}$ & $\mathbf{88.04 \pm 0.83}$ & $87.45 \pm 2.57$ & Moldova & $90.18 \pm 0.79$ & $\mathbf{91.25 \pm 0.84}$ & $86.89 \pm 3.01$ & $\mathbf{89.99 \pm 2.22}$\\
    India & $69.09 \pm 1.74$ & $\mathbf{71.94 \pm 2.06}$ & $\mathbf{80.04 \pm 3.24}$ & $79.73 \pm 3.65$ & Indonesia & $64.31 \pm 0.84$ & $\mathbf{65.98 \pm 0.97}$ & $70.28 \pm 1.64$ & $\mathbf{72.05 \pm 2.81}$\\
    Romania & $94.49 \pm 1.52$ & $\mathbf{96.01 \pm 0.73}$ & $91.65 \pm 1.21$ & $\mathbf{93.01 \pm 0.84}$ & Bermuda & $92.31 \pm 0.60$ & $\mathbf{92.82 \pm 0.68}$ & $93.70 \pm 0.35$ & $\mathbf{93.82 \pm 0.08}$\\
    Slovakia & $82.10 \pm 0.98$ & $\mathbf{84.16 \pm 2.60}$ & $90.31 \pm 3.88$ & $\mathbf{92.60 \pm 4.64}$ & Malaysia & $\mathbf{78.93 \pm 3.78}$ & $75.79 \pm 2.77$ & $\mathbf{94.16 \pm 0.49}$ & $89.06 \pm 5.22$\\
    Hungary & $\mathbf{48.92 \pm 3.59}$ & $48.48 \pm 4.55$ & $\mathbf{25.51 \pm 2.60}$ & $24.00 \pm 1.09$ & South Africa & $\mathbf{95.31 \pm 1.68}$ & $94.82 \pm 2.79$ & $\mathbf{96.87 \pm 0.96}$ & $96.83 \pm 2.11$\\
    Japan & $41.41 \pm 3.21$ & $\mathbf{42.63 \pm 4.13}$ & $35.33 \pm 1.28$ & $\mathbf{42.81 \pm 5.73}$ & Latvia & $\mathbf{93.66 \pm 0.64}$ & $93.04 \pm 0.85$ & $\mathbf{74.78 \pm 4.33}$ & $72.14 \pm 2.00$\\
    Iceland & $\mathbf{96.55 \pm 1.20}$ & $96.51 \pm 0.50$ & $\mathbf{97.38 \pm 1.18}$ & $96.77 \pm 0.43$ & Kazakhstan & $86.33 \pm 3.06$ & $\mathbf{89.52 \pm 4.82}$ & $\mathbf{94.12 \pm 1.94}$ & $92.60 \pm 5.40$\\
    Venezuela & $94.87 \pm 0.53$ & $\mathbf{95.37 \pm 0.45}$ & $93.05 \pm 2.02$ & $\mathbf{93.79 \pm 3.07}$ & New Caledonia & $99.48 \pm 0.15$ & $\mathbf{99.52 \pm 0.08}$ & $99.25 \pm 0.19$ & $\mathbf{99.38 \pm 0.34}$\\
    Philippines & $77.76 \pm 3.97$ & $\mathbf{82.87 \pm 5.17}$ & $\mathbf{81.95 \pm 8.07}$ & $81.30 \pm 3.50$ & Estonia & $87.08 \pm 1.89$ & $\mathbf{89.29 \pm 2.49}$ & $77.30 \pm 1.22$ & $\mathbf{80.24 \pm 5.49}$\\
    Slovenia & $95.37 \pm 0.23$ & $\mathbf{95.74 \pm 0.44}$ & $\mathbf{97.47 \pm 0.45}$ & $97.04 \pm 0.38$ & Singapore & $\mathbf{86.42 \pm 2.36}$ & $86.40 \pm 2.59$ & $86.87 \pm 2.01$ & $\mathbf{87.53 \pm 1.30}$\\
    Ukraine & $92.99 \pm 0.70$ & $\mathbf{93.45 \pm 0.76}$ & $92.60 \pm 1.84$ & $\mathbf{93.52 \pm 1.81}$ & Bangladesh & $\mathbf{78.61 \pm 0.43}$ & $76.48 \pm 1.06$ & $\mathbf{82.45 \pm 2.54}$ & $79.01 \pm 2.18$\\
    Belarus & $91.08 \pm 3.08$ & $\mathbf{94.43 \pm 3.59}$ & $\mathbf{96.40 \pm 1.76}$ & $94.44 \pm 4.19$ & Paraguay & $96.01 \pm 1.23$ & $\mathbf{96.68 \pm 0.32}$ & $\mathbf{97.20 \pm 0.35}$ & $96.25 \pm 1.17$\\
    Serbia & $95.31 \pm 0.48$ & $\mathbf{95.68 \pm 0.33}$ & $92.62 \pm 3.83$ & $\mathbf{93.09 \pm 1.81}$ & Cyprus & $\mathbf{97.67 \pm 0.34}$ & $97.57 \pm 0.31$ & $94.31 \pm 7.21$ & $\mathbf{98.32 \pm 0.37}$\\
    Croatia & $94.59 \pm 2.21$ & $\mathbf{96.40 \pm 0.57}$ & $88.04 \pm 4.68$ & $\mathbf{90.66 \pm 3.76}$ & Bosnia & $84.04 \pm 1.47$ & $\mathbf{87.42 \pm 1.95}$ & $84.46 \pm 5.76$ & $\mathbf{88.61 \pm 5.06}$\\
    Greece & $81.98 \pm 0.60$ & $\mathbf{85.00 \pm 1.61}$ & $40.97 \pm 14.89$ & $\mathbf{56.01 \pm 10.98}$ & Ireland & $87.44 \pm 0.69$ & $\mathbf{87.69 \pm 0.95}$ & $86.49 \pm 1.31$ & $\mathbf{87.56 \pm 3.01}$\\
    New Zealand & $94.27 \pm 1.50$ & $\mathbf{95.91 \pm 1.41}$ & $\mathbf{99.44 \pm 0.29}$ & $98.21 \pm 1.37$ & Algeria & $85.37 \pm 2.05$ & $\mathbf{86.03 \pm 1.55}$ & $84.65 \pm 4.47$ & $\mathbf{85.08 \pm 2.50}$\\
    Portugal & $93.65 \pm 0.46$ & $\mathbf{94.54 \pm 0.67}$ & $92.68 \pm 1.46$ & $\mathbf{93.33 \pm 0.59}$ & Colombia & $87.81 \pm 0.92$ & $\mathbf{89.09 \pm 0.64}$ & $\mathbf{89.51 \pm 0.88}$ & $88.32 \pm 2.55$\\
    Bulgaria & $\mathbf{91.03 \pm 2.07}$ & $90.87 \pm 2.63$ & $\mathbf{93.47 \pm 3.07}$ & $92.97 \pm 3.66$ & Uzbekistan & $86.76 \pm 1.13$ & $\mathbf{87.36 \pm 0.74}$ & $75.18 \pm 1.92$ & $\mathbf{77.52 \pm 2.62}$\\
    Lithuania & $87.67 \pm 3.05$ & $\mathbf{90.88 \pm 1.73}$ & $\mathbf{76.41 \pm 1.66}$ & $76.16 \pm 1.54$ &  && & &\\
    \bottomrule
    \end{tabular}%
    }
\end{table*}

\subsubsection{ADANN}
In a second attempt to improve our models' zero-shot performance, the base architecture was augmented with a domain adaptation approach as described in Subsection~\ref{subsec:adann}. The results of those two models (\fasttextadann{} and \bpembadann{}) are compared to their attention approach equivalents in \autoref{table:zero-shot-results}. For both the \fasttextadann{} and \bpembadann{} models (left and right section of the table respectively), we observe similar results. 

First, the ADANN algorithm was able to improve the performance for a minority (4 and 8 respectively for \fasttextadann{} and \bpembadann{} respectively) of the countries by more than $1~\%$. A few (2 or 3) out of the 41 improve by more than $2~\%$ for both models. Also, near a fourth of the 41 countries' accuracies increased by less than $1~\%$. For the other countries (the majority), we observe that for half of them, the difference is less than $1~\%$ poorer than the attention approaches, and the other half is a couple of percent poorer. Sometimes the difference can be as much as $5~\%$ (e.g. Sweden for \fasttextadann{} or Estonia for \bpembadann{}). Also, for both models, two countries drop below $80~\%$ (Lithuania and Moldova for \fasttextadann{} and Philippines and Bosnia for \bpembadann{}) and for \bpembadann{} Sweden accuracy's drops below $90~\%$. Thus, the use of domain adaptation technique during training lowers the performance overall, especially since only one country for each model (Malaysia and India respectively) yields results above $80~\%$ compared to the base approach and one above $90~\%$ (Kazakhstan and Malaysia respectively). Although, overall, the two models achieve good results for nearly $80~\%$ of the countries, almost $50~\%$ of which are near state-of-the-art performance. However, these results are less good than models using attention mechanisms in many cases.

Second, we observe that our model does not seem to have fit the training data as much as possible, as shown in \autoref{table:multinational-holdout-attention}. This table presents the multinational models' mean accuracy (and a standard deviation) on the holdout dataset for \fasttextadann{}. Nevertheless, we are surprised by our results since the ADANN algorithm is a transfer learning technique. An advantage of ADANN is that the network's weights have strong incentives to be subject-agnostic, meaning that the learned representation extracted from the network can be thought of as general features for the prediction layer \cite{adann}. We argue that it is more challenging to train models using the ADANN algorithm since the time needed for one epoch is nearly 5 hours. Meaning that the expected time to train for the 200 epochs is near 41 days per model (and we train five seeds) compared to a couple of days for the attention models. So we did not have much opportunity to fine-tune our model. We also hypothesize that the domain choice (i.e. the country) might be too granular since many countries have similar patterns or similar language, making the task more difficult. An idea of improvement could be to use more definitions for the domain, such as the language and the address pattern type. For example, we could use a categorical variable representing the address pattern number and a categorical variable for the language. That way, we could help guide the training into a better understanding of the context of an address which is not necessary the country but rather the language and the pattern. 

Finally, on average, performance is similar between the attention and ADANN approaches, but FastText models perform slightly better. They are, on average, $2~\%$ better than the BPEmb approaches. For example, \fasttextadann{} yields in average $86.45 \pm 12.13~\%$ accuracy across the zero-shot countries and \bpembadann{} yields a $85.32 \pm 15.31~\%$. Again, both BPEmb models have higher variance than FastText one. These results show that BPEmb models' performance tends to be more variable and more sensitive to changes in seeds.

\begin{table*}
    \centering
    \caption{Mean accuracy (and standard deviation) per country for zero-shot transfer models - attention models versus ADANN models.}
    \label{table:zero-shot-results-adann}
    \resizebox{\textwidth}{!}{%
    \begin{tabular}{lrr|rrlrr|rr}
    \toprule
    Country & FastTextAtt & FastTextADANN & BPEmbAtt & BPEmbADANN & Country & FastTextAtt & FastTextADANN & BPEmbAtt & BPEmbADANN\\
    \midrule
    Belgium & $90.01 \pm 1.59$ & $\mathbf{90.90 \pm 3.76}$ & $\mathbf{89.35 \pm 0.48}$ & $88.74 \pm 0.59$ & Faroe Islands & $\mathbf{74.00 \pm 1.00}$ & $69.76 \pm 2.08$ & $\mathbf{85.89 \pm 1.08}$ & $85.34 \pm 0.20$\\
    Sweden & $\mathbf{85.73 \pm 2.30}$ & $80.99 \pm 3.78$ & $\mathbf{90.64 \pm 5.07}$ & $89.65 \pm 1.08$ & Réunion & $\mathbf{96.86 \pm 1.27}$ & $96.48 \pm 0.96$ & $\mathbf{94.19 \pm 0.27}$ & $93.99 \pm 0.59$\\
    Argentina & $\mathbf{87.76 \pm 0.71}$ & $87.72 \pm 1.33$ & $\mathbf{87.45 \pm 2.57}$ & $87.13 \pm 2.65$ & Moldova & $\mathbf{91.25 \pm 0.84}$ & $89.87 \pm 0.56$ & $\mathbf{89.99 \pm 2.22}$ & $88.67 \pm 1.66$\\
    India & $\mathbf{71.94 \pm 2.06}$ & $70.30 \pm 1.22$ & $79.73 \pm 3.65$ & $\mathbf{81.12 \pm 2.97}$ & Indonesia & $\mathbf{65.98 \pm 0.97}$ & $65.42 \pm 0.75$ & $\mathbf{72.05 \pm 2.81}$ & $71.71 \pm 2.79$\\
    Romania & $96.01 \pm 0.73$ & $\mathbf{96.22 \pm 1.11}$ & $\mathbf{93.01 \pm 0.84}$ & $92.58 \pm 1.53$ & Bermuda & $\mathbf{92.82 \pm 0.68}$ & $92.02 \pm 0.52$ & $93.82 \pm 0.08$ & $\mathbf{95.02 \pm 0.58}$\\
    Slovakia & $\mathbf{84.16 \pm 2.60}$ & $81.93 \pm 1.80$ & $\mathbf{92.60 \pm 4.64}$ & $91.14 \pm 3.81$ & Malaysia & $75.79 \pm 2.77$ & $\mathbf{83.48 \pm 3.42}$ & $89.06 \pm 5.22$ & $\mathbf{90.52 \pm 4.27}$\\
    Hungary & $\mathbf{48.48 \pm 4.55}$ & $46.42 \pm 5.62$ & $\mathbf{24.00 \pm 1.09}$ & $23.54 \pm 1.69$ & South Africa & $\mathbf{94.82 \pm 2.79}$ & $93.09 \pm 2.35$ & $\mathbf{96.83 \pm 2.11}$ & $96.02 \pm 1.55$\\
    Japan & $42.63 \pm 4.13$ & $\mathbf{43.88 \pm 1.26}$ & $\mathbf{42.81 \pm 5.73}$ & $38.64 \pm 5.26$ & Latvia & $\mathbf{93.04 \pm 0.85}$ & $90.64 \pm 1.26$ & $\mathbf{72.14 \pm 2.00}$ & $71.92 \pm 2.60$\\
    Iceland & $\mathbf{96.51 \pm 0.50}$ & $96.18 \pm 0.35$ & $96.77 \pm 0.43$ & $\mathbf{97.15 \pm 0.64}$ & Kazakhstan & $89.52 \pm 4.82$ & $\mathbf{92.00 \pm 1.36}$ & $92.60 \pm 5.40$ & $\mathbf{96.49 \pm 1.94}$\\
    Venezuela & $\mathbf{95.37 \pm 0.45}$ & $95.02 \pm 0.76$ & $93.79 \pm 3.07$ & $\mathbf{94.79 \pm 0.57}$ & New Caledonia & $\mathbf{99.52 \pm 0.08}$ & $99.52 \pm 0.11$ & $99.38 \pm 0.34$ & $\mathbf{99.40 \pm 0.26}$\\
    Philippines & $82.87 \pm 5.17$ & $\mathbf{83.81 \pm 4.95}$ & $\mathbf{81.30 \pm 3.50}$ & $74.94 \pm 5.57$ & Estonia & $\mathbf{89.29 \pm 2.49}$ & $86.72 \pm 4.00$ & $80.24 \pm 5.49$ & $\mathbf{85.84 \pm 4.87}$\\
    Slovenia & $95.74 \pm 0.44$ & $\mathbf{95.75 \pm 0.53}$ & $\mathbf{97.04 \pm 0.38}$ & $96.77 \pm 0.57$ & Singapore & $\mathbf{86.40 \pm 2.59}$ & $84.96 \pm 2.84$ & $\mathbf{87.53 \pm 1.30}$ & $84.70 \pm 1.86$\\
    Ukraine & $93.45 \pm 0.76$ & $\mathbf{93.70 \pm 0.28}$ & $93.52 \pm 1.81$ & $\mathbf{94.17 \pm 2.30}$ & Bangladesh & $76.48 \pm 1.06$ & $\mathbf{76.80 \pm 1.21}$ & $\mathbf{79.01 \pm 2.18}$ & $78.68 \pm 3.96$\\
    Belarus & $94.43 \pm 3.59$ & $\mathbf{96.05 \pm 0.57}$ & $94.44 \pm 4.19$ & $\mathbf{98.15 \pm 0.97}$ & Paraguay & $96.68 \pm 0.32$ & $\mathbf{97.00 \pm 1.08}$ & $96.25 \pm 1.17$ & $\mathbf{96.28 \pm 0.96}$\\
    Serbia & $95.68 \pm 0.33$ & $\mathbf{95.75 \pm 0.25}$ & $93.09 \pm 1.81$ & $\mathbf{93.45 \pm 1.44}$ & Cyprus & $\mathbf{97.57 \pm 0.31}$ & $97.39 \pm 0.38$ & $\mathbf{98.32 \pm 0.37}$ & $97.53 \pm 1.49$\\
    Croatia & $\mathbf{96.40 \pm 0.57}$ & $94.98 \pm 2.01$ & $90.66 \pm 3.76$ & $\mathbf{91.37 \pm 4.39}$ & Bosnia & $\mathbf{87.42 \pm 1.95}$ & $84.95 \pm 3.47$ & $\mathbf{88.61 \pm 5.06}$ & $79.11 \pm 3.07$\\
    Greece & $\mathbf{85.00 \pm 1.61}$ & $83.18 \pm 3.31$ & $56.01 \pm 10.98$ & $\mathbf{57.47 \pm 7.75}$ & Ireland & $\mathbf{87.69 \pm 0.95}$ & $87.44 \pm 0.20$ & $87.56 \pm 3.01$ & $\mathbf{88.42 \pm 0.97}$\\
    New Zealand & $\mathbf{95.91 \pm 1.41}$ & $93.60 \pm 2.42$ & $98.21 \pm 1.37$ & $\mathbf{99.13 \pm 0.37}$ & Algeria & $\mathbf{86.03 \pm 1.55}$ & $83.07 \pm 3.85$ & $\mathbf{85.08 \pm 2.50}$ & $83.64 \pm 2.60$\\
    Portugal & $\mathbf{94.54 \pm 0.67}$ & $94.53 \pm 0.41$ & $\mathbf{93.33 \pm 0.59}$ & $91.07 \pm 1.26$ & Colombia & $\mathbf{89.09 \pm 0.64}$ & $87.76 \pm 1.30$ & $88.32 \pm 2.55$ & $\mathbf{88.99 \pm 1.83}$\\
    Bulgaria & $\mathbf{90.87 \pm 2.63}$ & $90.40 \pm 1.10$ & $92.97 \pm 3.66$ & $\mathbf{93.93 \pm 2.59}$ & Uzbekistan & $\mathbf{87.36 \pm 0.74}$ & $86.23 \pm 2.18$ & $\mathbf{77.52 \pm 2.62}$ & $73.83 \pm 2.92$\\
    Lithuania & $\mathbf{90.88 \pm 1.73}$ & $88.58 \pm 2.35$ & $76.16 \pm 1.54$ & $\mathbf{77.26 \pm 2.54}$ & &&&\\
    \bottomrule
    \end{tabular}%
    }
\end{table*}

\begin{table}
    \centering
    \caption{Mean accuracy (and standard deviation) for multinational models on holdout dataset for training countries - \fasttextatt{} versus \fasttextadann{}.}
    \label{table:multinational-holdout-attention}
    \resizebox{0.49\textwidth}{!}{%
    \begin{tabular}{lrrlrr}
    \toprule
    Country & FastTextAtt & FastTextADANN & Country & FastTextAtt & FastTextADANN\\
    \midrule
    United States & $\mathbf{99.73 \pm 0.02}$ & $99.70 \pm 0.03$ & Poland & $\mathbf{99.80 \pm 0.02}$ & $99.78 \pm 0.02$\\
    Brazil & $\mathbf{99.58 \pm 0.04}$ & $99.53 \pm 0.04$ & Norway & $99.44 \pm 0.11$ & $\mathbf{99.53 \pm 0.04}$\\
    South Korea & $\mathbf{99.98 \pm 0.01}$ & $\mathbf{99.98 \pm 0.01}$ & Austria & $\mathbf{99.38 \pm 0.06}$ & $99.30 \pm 0.07$\\
    Australia & $\mathbf{99.77 \pm 0.03}$ & $99.76 \pm 0.03$ & Finland & $\mathbf{99.83 \pm 0.02}$ & $99.82 \pm 0.01$\\
    Mexico & $\mathbf{99.71 \pm 0.03}$ & $99.68 \pm 0.02$ & Denmark & $\mathbf{99.82 \pm 0.03}$ & $99.80 \pm 0.02$\\
    Germany & $\mathbf{99.85 \pm 0.02}$ & $99.84 \pm 0.01$ & Czechia & $\mathbf{99.73 \pm 0.02}$ & $99.70 \pm 0.02$\\
    Spain & $\mathbf{99.83 \pm 0.02}$ & $99.80 \pm 0.02$ & Italy & $\mathbf{99.83 \pm 0.02}$ & $99.80 \pm 0.02$\\
    Netherlands & $\mathbf{99.75 \pm 0.03}$ & $99.72 \pm 0.02$ & France & $\mathbf{99.79 \pm 0.04}$ & $99.77 \pm 0.03$\\
    Canada & $\mathbf{99.87 \pm 0.02}$ & $99.85 \pm 0.01$ & UK & $\mathbf{99.77 \pm 0.05}$ & $99.73 \pm 0.03$\\
    Switzerland & $\mathbf{99.62 \pm 0.05}$ & $99.59 \pm 0.05$ & Russia & $\mathbf{99.40 \pm 0.13}$ & $99.23 \pm 0.12$\\
    \bottomrule
    \end{tabular}%
    }
\end{table}

\section{Missing Values Handling}
\label{sec:missingresults}
In this section, we aim to evaluate and improve the results of our four best models, \fasttextatt{}, \bpembatt{}, \fasttextadann{} and \bpembadann{} on the incomplete address data. As presented in Subsection~\ref{subsec:iad}, we introduce an incomplete address dataset where the addresses do not always include all the components. \autoref{table:incomplete-holdout} presents the results of the four models evaluated on the incomplete holdout dataset without any prior training on incomplete addresses. Since we observe that performances for all of the training countries are lower by $20~\%$ to $40~\%$ than previous scores, we choose to only evaluate our models on the countries seen during training (holdout). The lower accuracy is that of South Korea for both the models (nearly over the $45~\%$ accuracy)(will be discussed in more detail later). We also observe that the ADANN approach yields better results (12 out of 20) and is better by less than $1~\%$ on average (last row). Finally, we observe that the BPEmb approach still has higher variance than the FastText one since we have observed some seeds converging to suboptimal loss. Nevertheless, we observe that despite poorer performance on zero-shot evaluation, the ADANN approaches yield better results on incomplete addresses than attention approaches. We hypothesize that ADANN models have not overfitted the representation of an address structure seen in the training and more on the general features of address structures and domain type (e.g. the country and language). Also, knowing the domain, and indirectly the address structure and language, makes it easier to parse an address when incomplete.

\begin{table*}
    \centering
    \caption{Mean accuracy (and standard deviation) for multinational models without any prior training on incomplete dataset.}
    \label{table:incomplete-holdout}
        \resizebox{0.6\textwidth}{!}{%
    \begin{tabular}{lrrrr}
    \toprule
    Country & FastTextAtt & FastTextADANN & BPEmbAtt & BPEmbADANN\\
    \midrule
    United States & $\mathbf{68.59 \pm 2.61}$ & $65.57 \pm 0.80$ & $66.14 \pm 3.88$ & $\mathbf{67.94 \pm 4.94}$\\
    Brazil & $57.00 \pm 2.41$ & $\mathbf{57.39 \pm 3.63}$ & $\mathbf{55.01 \pm 3.90}$ & $51.10 \pm 4.08$\\
    South Korea & $45.91 \pm 4.22$ & $\mathbf{46.28 \pm 2.29}$ & $\mathbf{49.84 \pm 8.17}$ & $35.85 \pm 5.32$\\
    Australia & $75.90 \pm 1.28$ & $\mathbf{76.02 \pm 0.36}$ & $\mathbf{77.18 \pm 0.80}$ & $76.49 \pm 1.56$\\
    Mexico & $\mathbf{61.52 \pm 2.42}$ & $59.88 \pm 2.83$ & $\mathbf{61.25 \pm 1.42}$ & $58.93 \pm 6.36$\\
    Germany & $48.39 \pm 1.40$ & $\mathbf{50.32 \pm 6.96}$ & $\mathbf{58.98 \pm 0.86}$ & $58.62 \pm 3.31$\\
    Spain & $70.74 \pm 1.26$ & $\mathbf{72.39 \pm 1.13}$ & $79.51 \pm 0.49$ & $\mathbf{81.05 \pm 1.17}$\\
    Netherlands & $59.72 \pm 7.54$ & $\mathbf{63.74 \pm 3.47}$ & $75.34 \pm 1.32$ & $\mathbf{77.39 \pm 3.52}$\\
    Canada & $\mathbf{72.77 \pm 0.76}$ & $71.04 \pm 1.98$ & $73.00 \pm 2.51$ & $\mathbf{81.29 \pm 3.22}$\\
    Switzerland & $\mathbf{64.13 \pm 6.21}$ & $62.66 \pm 6.31$ & $\mathbf{73.08 \pm 0.89}$ & $72.55 \pm 3.22$\\
    Poland & $43.99 \pm 3.35$ & $\mathbf{47.04 \pm 4.06}$ & $48.50 \pm 1.08$ & $\mathbf{54.47 \pm 4.73}$\\
    Norway & $\mathbf{64.69 \pm 9.77}$ & $62.67 \pm 8.64$ & $76.43 \pm 1.62$ & $\mathbf{78.28 \pm 4.04}$\\
    Austria & $\mathbf{69.33 \pm 6.67}$ & $69.21 \pm 5.21$ & $77.69 \pm 1.08$ & $\mathbf{78.56 \pm 2.39}$\\
    Finland & $57.94 \pm 8.91$ & $\mathbf{60.83 \pm 6.98}$ & $74.65 \pm 1.36$ & $\mathbf{76.31 \pm 3.46}$\\
    Denmark & $\mathbf{56.94 \pm 3.24}$ & $55.54 \pm 5.48$ & $\mathbf{65.32 \pm 2.40}$ & $63.73 \pm 3.04$\\
    Czechia & $61.16 \pm 3.21$ & $\mathbf{64.40 \pm 2.91}$ & $72.58 \pm 1.72$ & $\mathbf{74.79 \pm 3.53}$\\
    Italy & $66.42 \pm 2.49$ & $\mathbf{71.16 \pm 1.67}$ & $73.89 \pm 0.77$ & $\mathbf{76.33 \pm 1.46}$\\
    France & $70.15 \pm 0.88$ & $\mathbf{73.14 \pm 2.38}$ & $71.91 \pm 3.46$ & $\mathbf{77.45 \pm 1.08}$\\
    UK & $53.63 \pm 1.28$ & $\mathbf{54.58 \pm 1.24}$ & $51.27 \pm 1.96$ & $\mathbf{53.62 \pm 3.30}$\\
    Russia & $\mathbf{58.41 \pm 1.12}$ & $57.94 \pm 0.99$ & $58.80 \pm 5.21$ & $\mathbf{61.91 \pm 2.86}$\\\midrule
    Mean  &       $61.37 \pm 8.65$&         $\mathbf{62.09 \pm 8.41}$ &    $67.02 \pm 9.91$ &      $\mathbf{67.83 \pm 12.15}$ \\
    \bottomrule
    \end{tabular}%
    }
\end{table*}

To improve our models' incomplete addresses performance, we have trained two new models using the complete and incomplete addresses datasets. This merged dataset consists of \num{150000} addresses per country using the same settings as described in \autoref{sec:exp}. We choose to only train the two best models on incomplete addresses, \fasttextadann{} and \bpembadann{}. We refer to these new models by \fasttextadannnoisy{} and \bpembadannnoisy{} where the difference between the two is the embeddings. \autoref{table:incomplete-holdout-train-clean-noisy} present the mean accuracy and a standard deviation of our two models tested on the incomplete dataset. 

First, we observe that using incomplete addresses during training substantially improves the accuracy for all the countries. We also observe that the \fasttextadannnoisy{} is the leading model across the board with the best accuracy on all the twenty countries, where results are nearly always more than $98~\%$ in accuracy. In contrast, we observe that, again, despite using an embeddings layer to learn the representation of the byte-pairs embeddings (BPEmb), \bpembadannnoisy{} shows poor results compare to \fasttextadannnoisy{}. Results are, on average, nearly $6~\%$ worse and have a higher variance of more than the double, with results as low as $84~\%$ (which is lower than some results observe in zero-shot evaluation). Again, this shows that the trained embeddings layer is overfitting or that the byte-pair embeddings are not well suited for address (such as the embeddings of postal code). This also could mean that the French fastText embeddings approach to construct out-of-vocabulary embeddings are more generalized than the one that we retrain using multi-lingual embeddings.

Second, it is interesting to note that even with a relatively large number of incomplete addresses in the training dataset (\num{50000}), we did not achieve scores as good as seen with the complete dataset (\autoref{table:multinational-holdout}). Results are in average of $98~\%$ and near $93~\%$ for \fasttextadann{} and \bpembadann{} respectively. Also, accuracies on the complete addresses holdout dataset are nearly as good as those presented in \autoref{table:multinational-holdout}. We observe similar results for most of the training countries with a difference of $0.5~\%$ lower.

Finally, we can see that the worse results for both models are for South Korea. These results contrast with the nearly perfect score observed for all the models in \autoref{table:multinational-holdout}. This highlights our hypothesis (Subsection~\ref{subsec:multieval}) that our model has memorized the particular pattern of South Korea during training for complete addresses. However, since we also have trained using incomplete addresses, some incomplete addresses are now not so different from the other address patterns, confusing our models. For example, if we remove the tags Province and Municipality of an address, it can be in 4 of the five patterns making more harder for our models. This shows that our models have overfitted in that case, and adding noise in the training data helps reduce that overfitting but lowers the accuracy. We also observe that using a domain adversarial technique substantially improves performance for that specific case where we observe the best improvement with the accuracy nearly doubling.

\begin{table*}
    \centering
    \caption{Mean accuracy (and standard deviation) for multinational models trained on complete and incomplete datasets and evaluated on the incomplete dataset for training countries. \textbf{bold} values indicate the best accuracy for each country.}
    \label{table:incomplete-holdout-train-clean-noisy}
    \resizebox{0.65\textwidth}{!}{%
    \begin{tabular}{lrrrr}
    \toprule
    Country & FastTextADANN & FastTextADANNNoisy & BPEmbADANN & BPEmbADANNNoisy\\
    \midrule
    United States & $65.57 \pm 0.80$ & $\mathbf{97.88 \pm 2.04}$ & $67.94 \pm 4.94$ & $92.12 \pm 3.60$\\
    Brazil & $57.39 \pm 3.63$ & $\mathbf{98.39 \pm 1.90}$ & $51.10 \pm 4.08$ & $89.57 \pm 6.46$\\
    South Korea & $46.28 \pm 2.29$ & $\mathbf{92.25 \pm 3.72}$ & $35.85 \pm 5.32$ & $84.58 \pm 9.86$\\
    Australia & $76.02 \pm 0.36$ & $\mathbf{98.86 \pm 1.46}$ & $76.49 \pm 1.56$ & $93.32 \pm 5.75$\\
    Mexico & $59.88 \pm 2.83$ & $\mathbf{97.56 \pm 1.63}$ & $58.93 \pm 6.36$ & $88.79 \pm 5.12$\\
    Germany & $50.32 \pm 6.96$ & $\mathbf{99.20 \pm 0.74}$ & $58.62 \pm 3.31$ & $96.79 \pm 2.05$\\
    Spain & $72.39 \pm 1.13$ & $\mathbf{98.24 \pm 1.82}$ & $81.05 \pm 1.17$ & $90.23 \pm 7.02$\\
    Netherlands & $63.74 \pm 3.47$ & $\mathbf{98.65 \pm 0.98}$ & $77.39 \pm 3.52$ & $97.21 \pm 1.77$\\
    Canada & $71.04 \pm 1.98$ & $\mathbf{97.88 \pm 2.65}$ & $81.29 \pm 3.22$ & $91.93 \pm 3.74$\\
    Switzerland & $62.66 \pm 6.31$ & $\mathbf{99.03 \pm 0.78}$ & $72.55 \pm 3.22$ & $97.12 \pm 1.52$\\
    Poland & $47.04 \pm 4.06$ & $\mathbf{99.28 \pm 0.52}$ & $54.47 \pm 4.73$ & $97.89 \pm 1.80$\\
    Norway & $62.67 \pm 8.64$ & $\mathbf{99.35 \pm 0.54}$ & $78.28 \pm 4.04$ & $98.25 \pm 1.06$\\
    Austria & $69.21 \pm 5.21$ & $\mathbf{99.10 \pm 1.16}$ & $78.56 \pm 2.39$ & $94.74 \pm 2.88$\\
    Finland & $60.83 \pm 6.98$ & $\mathbf{98.97 \pm 0.64}$ & $76.31 \pm 3.46$ & $98.75 \pm 0.57$\\
    Denmark & $55.54 \pm 5.48$ & $\mathbf{97.38 \pm 2.56}$ & $63.73 \pm 3.04$ & $92.15 \pm 3.77$\\
    Czechia & $64.40 \pm 2.91$ & $\mathbf{98.14 \pm 1.74}$ & $74.79 \pm 3.53$ & $93.75 \pm 3.79$\\
    Italy & $71.16 \pm 1.67$ & $\mathbf{98.82 \pm 1.15}$ & $76.33 \pm 1.46$ & $94.49 \pm 3.72$\\
    France & $73.14 \pm 2.38$ & $\mathbf{98.98 \pm 1.32}$ & $77.45 \pm 1.08$ & $91.10 \pm 6.85$\\
    UK & $54.58 \pm 1.24$ & $\mathbf{96.01 \pm 4.75}$ & $53.62 \pm 3.30$ & $84.95 \pm 7.63$\\
    Russia & $57.94 \pm 0.99$ & $\mathbf{96.05 \pm 4.05}$ & $61.91 \pm 2.86$ & $86.71 \pm 5.71$\\\midrule
    Mean             &         $62.09 \pm 8.41$ &              $\mathbf{98.00 \pm 1.62}$ &      $67.83 \pm 12.15$ &           $92.72 \pm 4.22$ \\
    \bottomrule
    \end{tabular}%
    }
\end{table*}

\section{Conclusion}
\label{sec:conclusion}
We estimate that we have reached our first objective, which was to build a single model capable of learning to parse addresses of different formats and languages using a multinational dataset and subword embeddings. Indeed, all our approaches achieve accuracies around $99~\%$ on all the twenty countries used for training. Our experiments with zero-shot transfer learning also yielded interesting results. First, our baseline approaches obtain good results achieving near $50~\%$ of state-of-the-art performance. Second, using an attention mechanism helps to improve our results and could also provide insights about the address elements on which the model focuses to make a tag prediction. However, this analysis is left as future work. Third, our experiments indicate that using a domain adversarial training algorithm does not necessarily improve our results on countries not seen during training. However, they bring a significant contribution on incomplete addresses. Finally, we tested some of our models on incomplete addresses to evaluate their performance on such data. We also improved performance by using some incomplete addresses during training improved performance. These results provide insights into the direction that our future work should take. It would be interesting to explore how other subword embeddings techniques, such as character-based ones \cite{kim2015characteraware}, would perform on the multinational address parsing task. Additional qualitative analysis of the results would also be required to investigate the models' typical errors further. 

\section*{Acknowledgment}
This research was supported by the Natural Sciences and Engineering Research Council of Canada (IRCPJ 529529-17) and a Canadian insurance company. We wish to thank the reviewers for their comments regarding our work and methodology.

\bibliographystyle{IEEEtran}
\bibliography{MAPZSE}

\end{document}